\documentclass[conference]{IEEEtran}
\usepackage{blindtext, graphicx}
\usepackage{array}
\usepackage{subfigure}
\usepackage{booktabs}
\usepackage{multirow}
\usepackage{amsmath}
\usepackage{textcomp}
\usepackage{array}

\ifCLASSINFOpdf

\else

\fi

\begin{document}

\title{Convolutional Neural Networks for Aerial Multi-Label Pedestrian Detection}

\author{\IEEEauthorblockN{Amir Soleimani,
Nasser M. Nasrabadi}
\IEEEauthorblockA{Lane Department of Computer Science and Electrical Engineering, West Virginia University\\ 
Email: a.soleimani.b@gmail.com,
nasser.nasrabadi@mail.wvu.edu}\\
This paper has been accepted in the 21st International Conference on Information Fusion and would be indexed in IEEE}

\maketitle

\begin{abstract}
The low resolution of objects of interest in aerial images makes pedestrian detection and action detection extremely challenging tasks. Furthermore, using deep convolutional neural networks to process large images can be demanding in terms of computational requirements. In order to alleviate these challenges, we propose a two-step, yes and no question answering framework to find specific individuals doing one or multiple specific actions in aerial images. First, a deep object detector, Single Shot Multibox Detector (SSD), is used to generate object proposals from small aerial images. Second, another deep network, is used to learn a latent common sub-space which associates the high resolution aerial imagery and the pedestrian action labels that are provided by the human-based sources.
\end{abstract}


%
\IEEEpeerreviewmaketitle

\section{Introduction}
Pedestrian detection is a task that has several applications from driver assistance systems and surveillance to image and video understanding. There has been a significant research on pedestrian detection since the late 1990s \cite{oren1997pedestrian} and more recently using deep learning \cite{tome2016deep} and \cite{zhang2016faster}. However, to the best of our knowledge, almost all works in this area have been limited to the frontal views from personal and closed-circuit television cameras (CCTVs). In contrast, detecting pedestrians in top-down or aerial
imagery is a relatively new area of research. Aerial pedestrian detection has several other applications like surveillance using Unmanned Aerial Vehicle (UAV), which  provides a wider range of view,  higher performance, search and rescue tasks, and human interaction understanding.\\
In this paper, we investigate the problem of finding particular pedestrians who are doing one or multiple specific actions (e.g., calling and running) in aerial images (see Figure 1). This line of research is particularly applicable in scenarios where the location of an individual exhibiting a malbehavior needs to be identified. For instance, in an urgent situation, on the basis of people\textquotesingle s tweets, it might be very important to search for a person who is running and carrying something in a street. In this paper, we assume that objects of interest can be discriminated based on their single or multiple actions, and evaluate the proposed framework on the Okutama-Action dataset \cite{okutama}, which is an aerial dataset for concurrent human single and multiple action detection.\\
\begin{figure}[!t]
\centering
\includegraphics[scale=0.06]{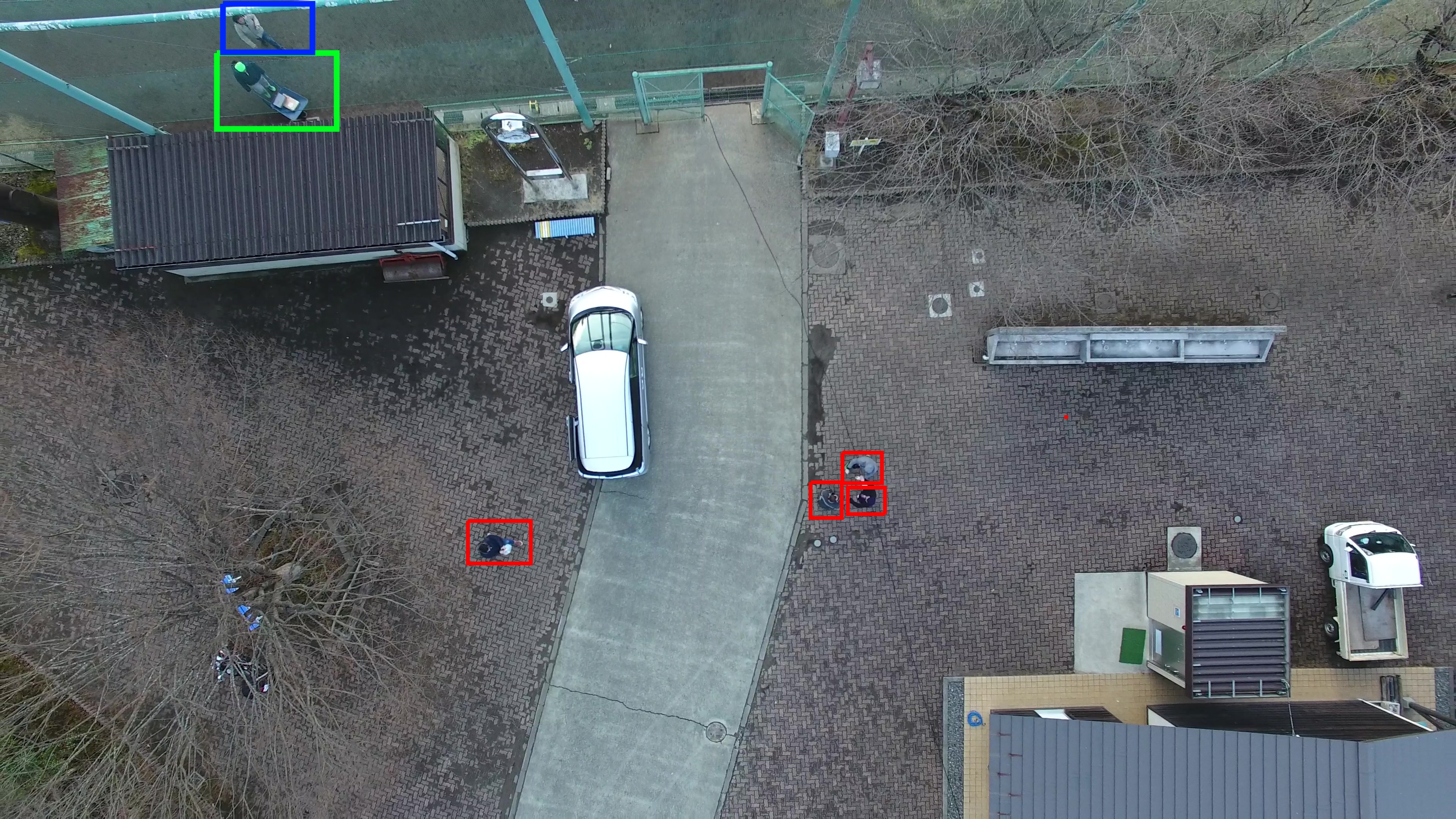}
\caption{Who is walking? Who is walking while pushing a wheelbarrow? Blue and Green boxes are respectively the targets for the first and second question, while Red boxes are the wrong answers. Image is a frame from the Okutama-Action dataset \cite{okutama}.}
\label{fig_graph}
\end{figure}
Although aerial imagery provides a wider range of view for surveillance tasks, it brings different challenges in the literature. In top-down view, pedestrians as well as other objects are very small. In addition, the state of the art object detectors like Faster R-CNN \cite{faster}, YOLO \cite{yolo}, SSD \cite{ssd} as well as deep object classifiers like VGG \cite{vgg} usually work with input images of less than one mega pixel (e.g., 300x300 and 500x500), which makes the problem even worse. In other words, pedestrians and objects of interest sizes are limited to a few pixels and this makes the recognition of single and multiple actions impossible. Figure 2 shows how pedestrian's appearance fade when resolutions decreases. \\
In this paper, we propose a two-step framework to generate objects of interest or object proposals from small size aerial images in the first step, and then we create a latent common sub-space and fuse high resolution object proposals with different possible actions in the second step. The final output of the framework is Yes or No for the question of "Is this probable pedestrian doing action X (and Y)?". The proposed system helps to find particular targets in aerial images.\\
This paper is organized as follows: In Section II, we review the literature of deep object detectors, multi-label classification, and visual question answering. Then in Section III, we explain our proposed framework. Section IV describes the experiments, their details, and results. Finally, in section V we summarize our conclusions.
\begin{figure}[!t]
\centering
\includegraphics[scale=0.322]{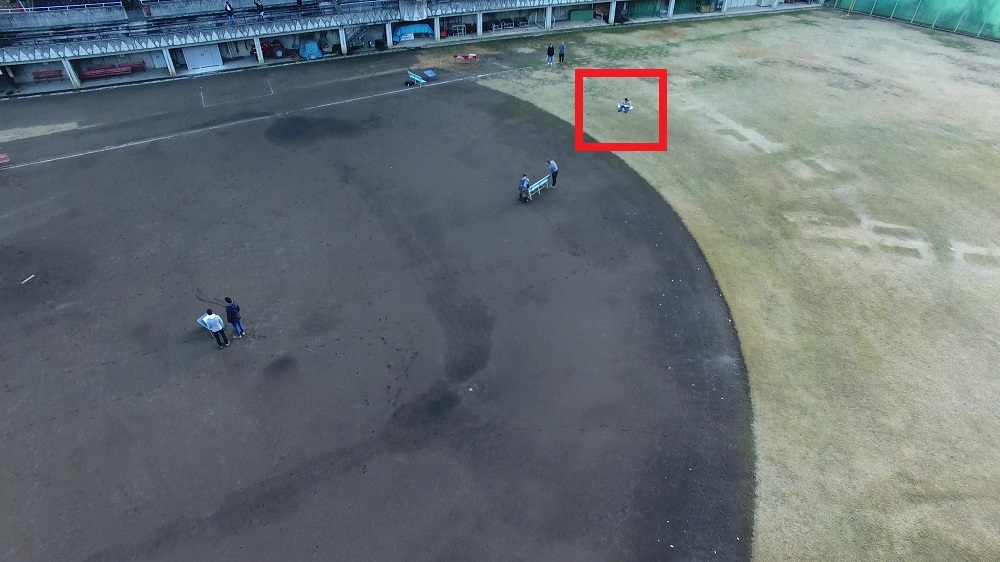}\\
\vspace{1mm}
\begin{tabular}{cc}
 \includegraphics[scale=1]{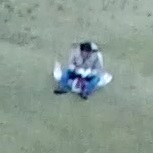}& \includegraphics[scale=6.95]{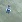}
\end{tabular}
\caption{Effect of resolution on aerial images. The left bottom patch is in the original resolution while the right one was cropped from the 300x300 image. Image is a frame from the Okutama-Action dataset \cite{okutama}.}
\label{fig_graph2}
\end{figure}

\section{Related Work}
\subsection{Deep Object Detector}
State of the art object detectors are divided into two categories. In the first category, input images in conjunction with thousands object proposals created by a method like selective search \cite{uijlings2013selective} are given to a network, and the network resamples pixels or extract features from these proposals, and finally a classifier determines if there is an object (in our case pedestrian) in the proposals or image patches. R-CNN \cite{rcnn} and Fast R-CNN \cite{fastrcnn} are the well-known deep detectors in this category. Faster R-CNN \cite{faster} is also another similar method, but in this network, object proposals for the first time in the literature are built using a deep structure. Faster R-CNN then uses Fast R-CNN in order to classify image proposals.\\
In the second category, the system is end-to-end that means only images (without object proposals) are given to a network and this network tries to both classify and localize objects. Overfeat \cite{overfeat}, YOLO \cite{yolo}, and SSD \cite{ssd} are the most promising systems in this category. In the first category, since the proposal making phase and classification phase are done separately, the error in the classification phase cannot propagate into the first phase and consequently cannot enhance the performance of the first phase. However, the end-to-end nature of these algorithms results in both the increase in speed and accuracy.\\
Among the deep detectors, SSD has outperformed other methods, because it detects objects in a multi-scale framework. First layers of the SSD network are made of a deep convolutional network (e.g., VGG), and then a number of extra layers are added to the base network in which the first layers try to detect smaller objects while the final layers focus on the larger objects.\\
In aerial pedestrian detection, objects are significantly smaller than the typical objects in well-known datasets like VOC \cite{everingham2010pascal} and ImageNet \cite{deng2009imagenet}. Moreover, pedestrian sizes can change significantly because of the changes in the drone flying altitude or the subject actions (e.g., standing or lying). In this paper, we focus on SSD because of these two facts and the promising results of SSD on other datasets. In the following subsection, we shortly describe SSD.
\subsection{SSD}
SSD \cite{ssd} is a feed-forward convolutional network that produces a collection of bounding boxes and corresponding scores showing the possibility of presence of an object in these boxes. In the standard SSD (SSD300), 8732 bounding boxes and 8732 scores per class are produced for each input image. The outputs are then followed by a non-maximum suppression step to produce the final detections. Non-maximum suppression is an algorithm that tries to eliminate the extra bounding boxes based on their scores and their intersections. In other words, non-maximum suppression tries to merge all the bounding boxes proposed for unique objects.\\
In the SSD's early layers a standard architecture, called base network, is used for high quality image representation (truncated before any classification layers). Authors then add auxiliary layers to the base network to produce detections. To build a multi-scale framework, these layers decrease in size, progressively.\\
SSD considers a number of default boxes separately for each extra layer or for a selection of these extra layers. These default boxes have different scales and aspect ratios, which are responsible for objects at different scales and aspect ratios. The idea behind the default boxes is to force the network to learn different object detectors based on the probable scale and aspect ratio of the objects. The sizes of feature maps produced by SSD's extra layers are decreased step-by-step, and hence, larger default boxes are assigned to the final layers. A number of 3-by-3 convolutional filters are applied to each or almost each extra layer of SSD.\\
Equation (1) is the training objective function for SSD. It consists of a confidence loss (softmax loss) which shows the degree of certainty for the existence of an object in the produced bounding boxes, and localization loss which is a smooth L1 loss and shows the error between the ground-truth and the produced bounding boxes. $x$ is a parameter that determines if there is an object in a default box or not, using the ground-truth coordinates and default boxes. $c$ is the class parameter, and $l$ and $g$ are used for produced locations and ground-truth locations, respectively \\
\begin{equation}
l(x,c,l,g)=1/N(L_{conf}(x,c)+{L_{loc}(x,l,g)}).
\end{equation}
\begin{figure*}[!t]
\centering
\includegraphics[scale=1]{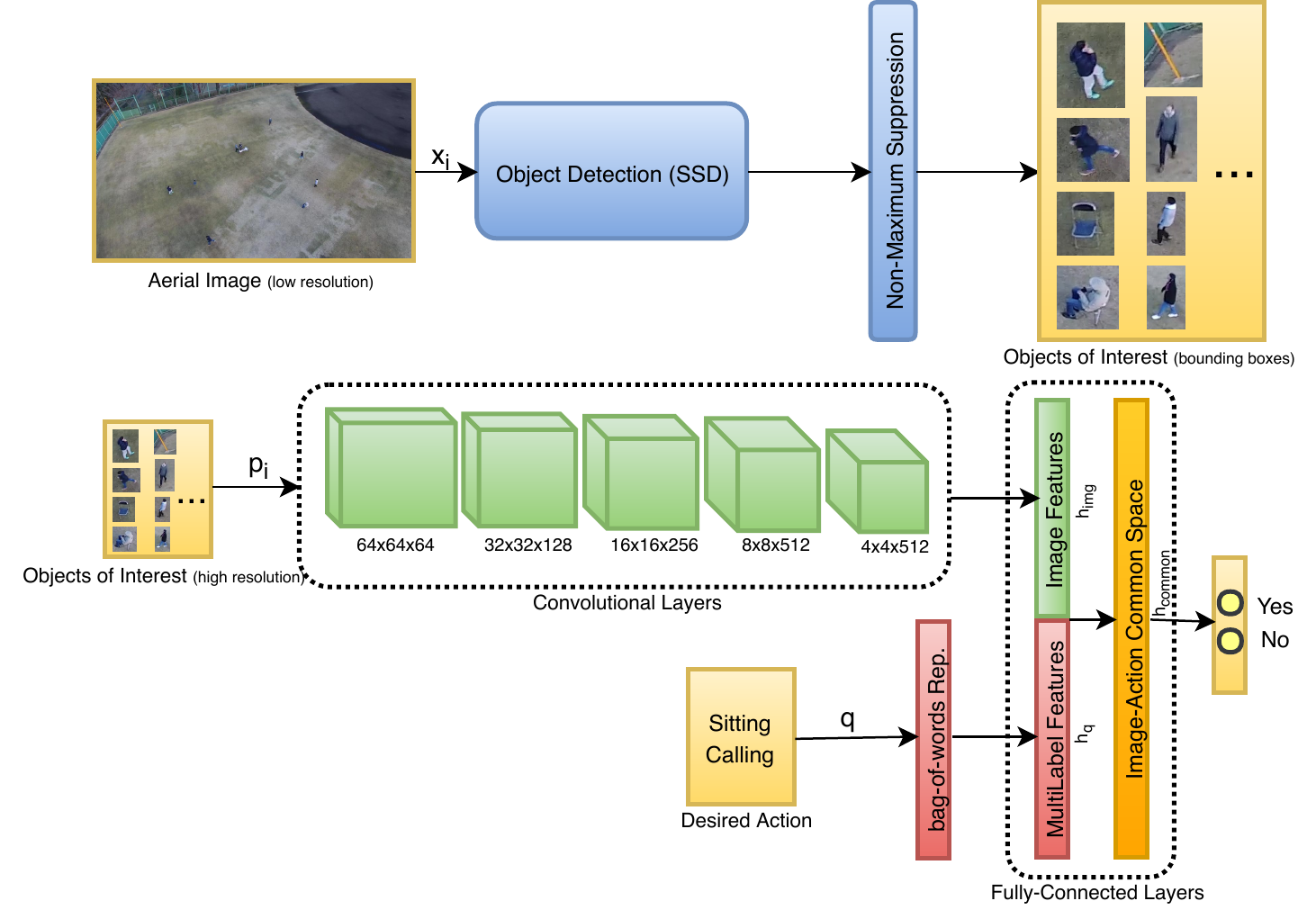}\\
\caption{SSD (blue) generates k objects of interest for each small-size aerial image, then a VGG network (green) extracts visual features from these objects of interest in their original sizes. Query or desired action(s) is then converted to a bag-of-words representation and pass through a fully-connected layer (red). Visual and textual information are fused in the next fully-connected layer (orange), and finally, a softmax classifier determines if the object of interest includes an individual doing the desired action or not.}
\label{fig_graph3}
\end{figure*}
\subsection{Multi-Label Classification}
Multi-label classification is a type of classification problem in which instead of assigning only one label to each input, inputs have more than one label. In our application, it means that a subject can do more than one action simultaneously (e.g., sitting and reading, standing and reading). The goal in multi-label classification is to learn the dependencies among different labels.\\
Binary Relevance (BR) is the baseline approach for transforming multi-label problem into a single-label problem \cite{Read2011}. However, this approach merely makes separate binary problems, and cannot model label dependencies. \\
The most straightforward method to make a multi-label classification in a neural network architecture is to use the sigmoid activation function instead of the softmax activation function, and use the value of each neuron in the output layer as scores for the corresponding label. However, this kind of architecture does not have the ability to answer to unseen and open-ended queries or questions. To this end, a reasonable approach is to design a common latent sub-space for images and textual or label information, and have a Visual Question Answering (VQA) system.
\subsection{Visual Question Answering}
Visual question answering is a new application in the field of information fusion. The most well-known work in this area is the VQA paper and its associated dataset \cite{vqa}. Authors in this paper, investigate the task of open-ended answering to questions created based on images. They also provide a dataset containing 0.25M images, 0.76M questions, and 10M answers.\\
For their baseline methods, they use two different neural network architectures. The first one is a multi-layer perception (MLP) which fuses the textual and visual information. For the textual information they use a bag-of-words representation from the top 1000 words in the VQA dataset's questions, and for the visual part they merely use the 4096 features created by the last hidden layer of their VGG network. Their second method is a LSTM model followed by a softmax layer to generate answers.

\section{Proposed Method}
The state of the art deep detectors like SSD, due to the limited computational resources, cannot be trained on high resolution images, therefore, the aerial action detection seems impossible using these types of networks. Authors in \cite{okutama} show that the SSD512 can detect pedestrian (not their actions) with the good accuracy of Mean Average Precision (mAP@0.5) of 72.3\% which is as good as the performance of SSD on the frontal-view datasets like VOC2017 \cite{ssd}. However, using even a larger input of 960x540, they only report mAP=18.18\% for detecting pedestrians' actions.\\
We propose a two-step framework (Figure 3) that in the first step, a SSD network generates objects of interest or object proposals using the small-sized aerial images, separately for each frame. In the second step, a yes and no visual question answering system determines the existence of a desired single or multiple actions.\\
Consider $X=\{x_1,x_2,...,x_n\}$ as a sequence of aerial images. For frame $x_i$, the SSD acts as a function $f(x_i)$ that generates $m_i$ bounding boxes for which we recall their corresponding image patches in the original high-resolution aerial image as object proposals $P_i=\{p_{i1},p_{i2},...,p_{im_i}\}$. Note that a simple threshold is applied on the confidence scores of the generated bounding boxes and then the non-maximum suppression removes the high overlapping bounding boxes. The SSD is trained in a way to generate similar bounding boxes to the ground-truth in terms of the Jaccard Index or Intersection Over Union (IoU), which is defined by the area of overlap between the ground-truth and predicted bounding boxes, divided by the area of the union of the two. A VGG network extracts features $h_{img}$ from each proposal $p_{ij}$ using the function $g_{img}(p_{ij})=h_{{img}_{ij}}$ consisting of several convolutional and fully-connected layers. The desired action $q$ is represented by using the bag-of-words representation which is transformed to $h_q$ using another fully-connected layer represented by function $g_q(q)=h_q$. We concatenate $h_q$ and $h_{img}$ and transform the resulting vector into a latent common sub-space using $g_{shared}(h_{img},h_q)=h_{common}$. Finally, a softmax classifier with two outputs decides about the existence of action(s) $q$ in proposal $p_{ij}$.\\
We use the SSD512 for the first step of the framework. A confidence threshold of 0.01 and NMS with the Jaccard Index of 0.45 is applied to remove extra proposals, and finally we keep 50 proposals per frame. This is a very fast step and done with speed of about 19FPS \cite{ssd}.\\
For the second step, we use the VGG16 \cite{vgg} structure with exception of only one fully-connected with 1000 neurons to create 1000 dimensional image features. Input images are re-scaled to 128x128. Desired action(s) are converted to 13 dimensional bag-of-words representation, and next mapped to 100 dimensional multi-label features using a fully-connected layer. Image and multi-label features are concatenated and transformed into 100 dimensional image-action common sub-space using another fully-connected layer.\\
\section{Experiments}
\subsection{Dataset}
Okutama-Action\cite{okutama} is an aerial view concurrent human action detection dataset. It contains 43 fully-annotated sequences of 12 human action classes. They used two UAV pilots to capture 22 different scenarios using 9 participants from a distance of 10-45 meters and camera angles of 45-90. The dataset is in the form of 30 FPS videos that provides 77365 images. They splitted the dataset into the training (33 videos) and testing samples (10 videos). Figure 1, shows a frame from the Okutama-Action dataset. They used the Amazon Mechanical Turk to manually annotate the frames in terms of bounding boxes and actions. Bounding boxes corresponding to pedestrians have one or more than one action label since subjects acted multiple actions simultaneously. Action labels are divided into three categories of human to human interactions (handshaking, hugging), human to object interactions (reading, drinking, pushing/pulling, carrying, calling), and none-interaction (running, walking, lying, sitting, standing). Single actions can be one of these 13 specific actions, and multiple actions almost consist of one none-interaction action and one action from the other two categories.\\
In this paper, we experimented on the Okutama-Action dataset \cite{okutama}. Its training set consists of 33 videos that provides about 60k frames. The authors have not published ground-truth for the test set, and therefore, we split the training set into two independent sets. We selected 24 videos for the training and isolated 9 remaining videos for the test phase. Test set consists of the following video files: 1.1.7, 1.1.10, 1.1.11, 1.2.9, 1.2.11, 2.1.7, 2.1.10, 2.2.9, 2.2.11. Table I shows the details about our defined training and test sets. Note that the lost and occluded objects have not been considered in neither of the sets. 
\begin{table}[h]
\centering
\begin{tabular}{ | c | c| c| c |  } 
\hline
Set & \#Frames & \#Objects & Object/Frame \\ 
\hline
Training & 39335 & 240520 & 6.11 \\ 
Test & 14898 & 75915 & 5.09\\ 
\hline
\end{tabular}
\caption{Training and Test sets details}
\label{table:2}
\end{table}
\subsection{Experimental Results}
We use Mean Average Precision (mAP) to show the results of pure pedestrian detection (without providing the query actions) using SSD300 and SSD512. We used the SSD unofficial pytorch code\footnote{https://github.com/amdegroot/ssd.pytorch}. The initial values for the base network has been replaced by VGG-16 weights trained on the ImageNet dataset and other initialization are produced randomly using the Xavier method \cite{xavier}. We set batch size of 16, and train SSD for 120k iterations. Other details of the network are similar to the original SSD paper \cite{ssd}, except augmentation that we only used random sample cropping and random mirroring.\\ 
Table II shows the results on the test set according to different Jaccard Indexes. Considering the low resolution of objects in aerial images and the performance of the state of the art detectors including SSD on other datasets like VOC2007, it is clear that SSD300 and SSD512 can obtain fair results for detecting pedestrians. For the next step of our framework, we use SSD512. Note that the higher accuracy of 72.3\% reported in \cite{okutama} might be because of a larger training set (our training set is divided into training and test sets).\\
Figure 4 shows the graphical results of SSD512 on two frames from the test set. As Table II and Figure 4 show, some errors occur in the first step, and can be propagated to the second step of the framework. Therefore, we use more predicted bounding boxes or proposals per image to cover probable objects better. Proposals can be pure background that can be recognized in the second step.

\begin{figure}[!t]
\centering
\begin{tabular}{c}
\includegraphics[scale=0.1815]{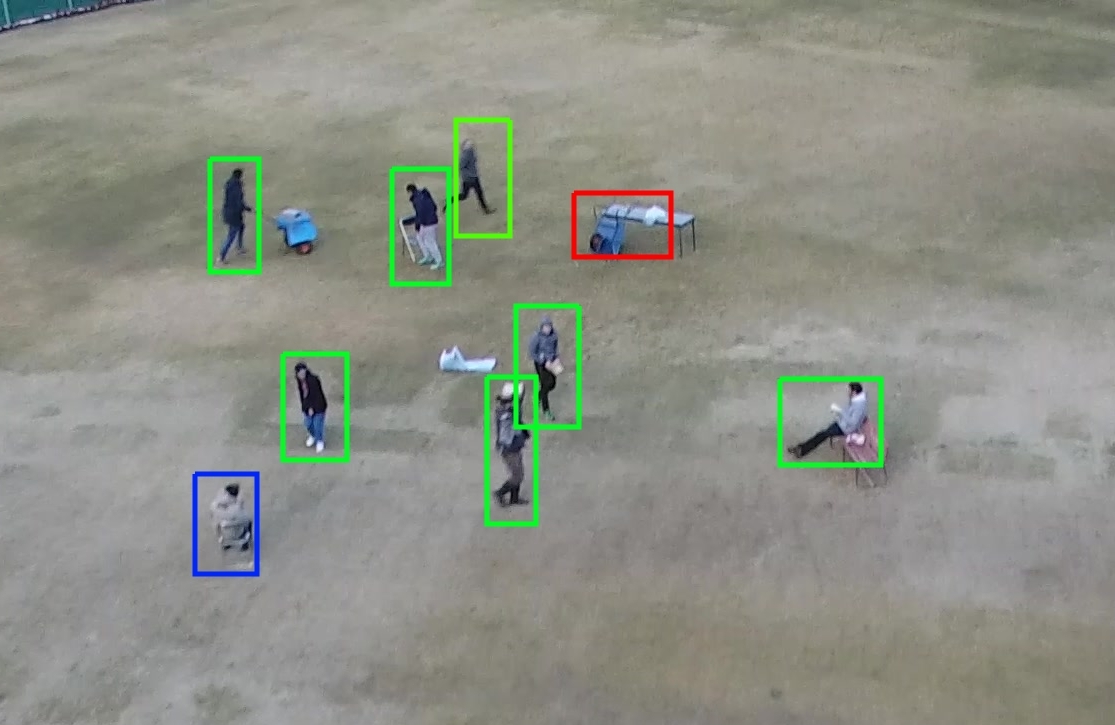}\\
\includegraphics[scale=0.138]{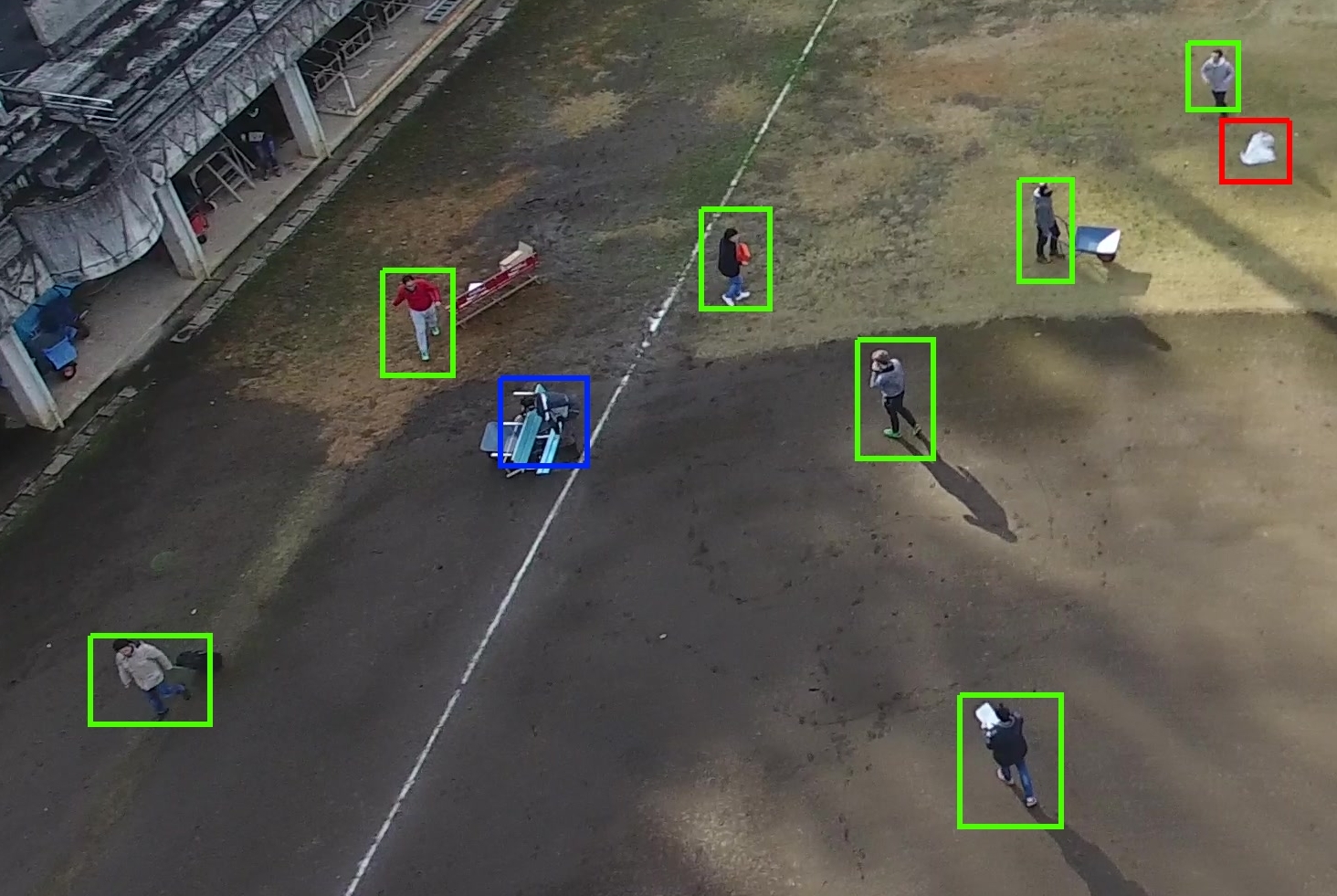}
\end{tabular}
\caption{Predicted bounding boxes on the test Set. Green, Blue, and Red are respectively for true, missed, and wrong predictions. Images were cropped for better view.}
\label{fig_graph4}
\end{figure}

\begin{table}[t]
\centering
\begin{tabular}{ | c |  c| c| } 
\hline
mAP & SSD300 & SSD512\\ 
\hline
mAP@0.50 & 37.8\%& 58.4\%\\
mAP@0.45 & 45.0\%& 65.8\%\\
mAP@0.40 & 59.0\%& 70.1\%\\ 
mAP@0.35 & 67.1\%& 75.3\%\\ 
mAP@0.30 & 72.5\%& 85.8\%\\ 
\hline
\end{tabular}
\caption{Pedestrian Detection Results for different Jaccard Indexes}
\label{table:3}
\end{table}

Figure 5 shows the comparison between the proposed method (two-step framework) and results of only SSD512 and SSD960 for action detection reported in \cite{okutama}. Note that although we trained the model on multiple action dataset, to provide comparability we only reported single action detection results in this figure. In all actions, the performance of the proposed method surpasses those reported using only SSD, and as expected this is because of low resolution of input images for detecting actions as well as using only single label information.\\
Table III compares the performance of using only SSD for detecting actions in the Okutama-Action dataset and the performance of the proposed two-step framework. We see that the proposed method even surpasses the results of SSD960 and again this is because the proposed method exploits higher resolution (the original high resolution of proposals in original images) and multi-action information.
\begin{figure}[!t]
\centering
\includegraphics[scale=0.22]{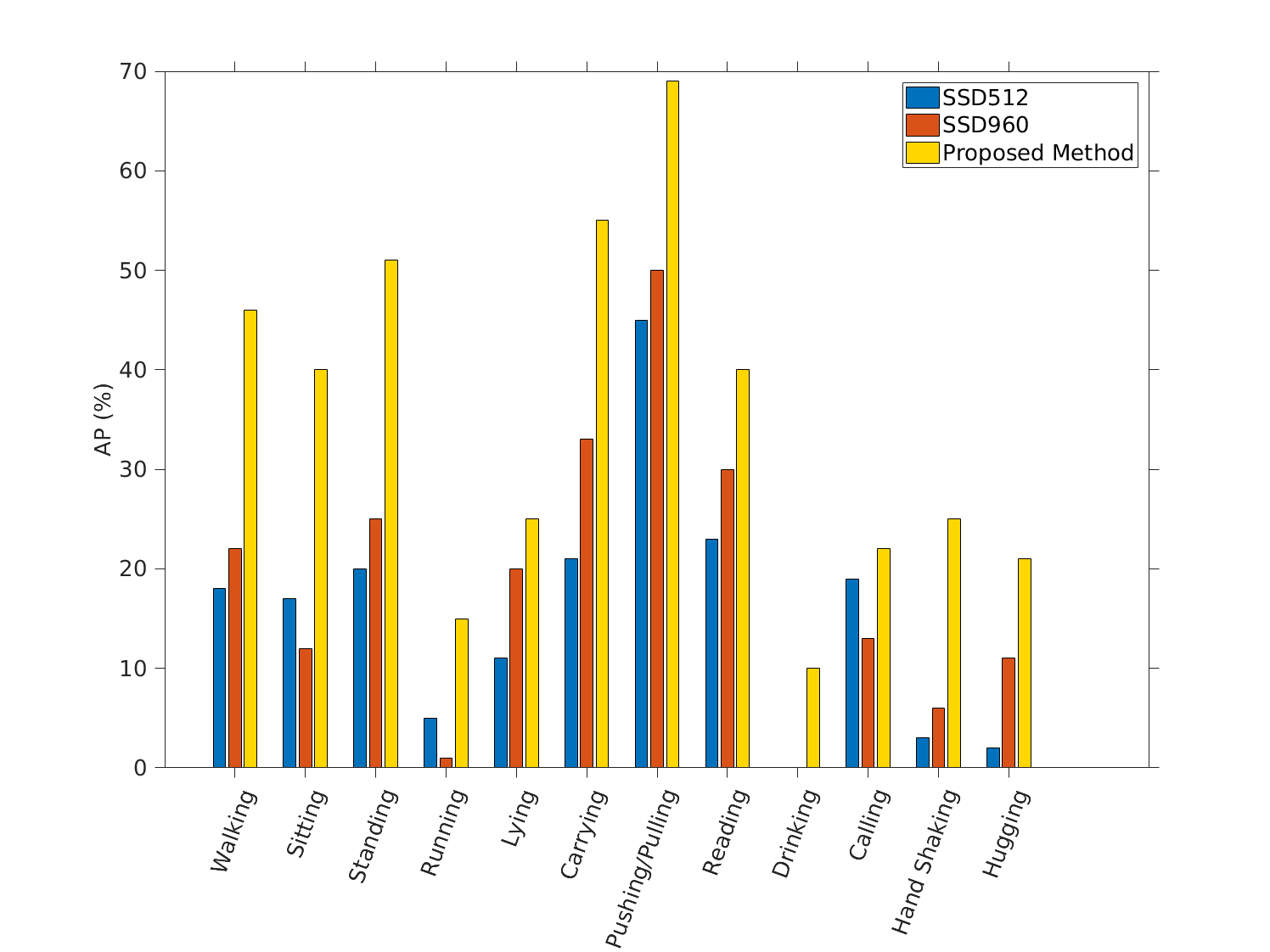}
\caption{Comparison between the proposed method and using only SSD for action detection.}
\label{fig_graph5}
\end{figure}

\begin{table}[t]
\centering
\begin{tabular}{ | c |  c| } 
\hline
Method & mAP@0.5 \\ 
\hline
SSD512 \cite{okutama}& 15.39\%\\\
SSD960 \cite{okutama} & 18.80\%\\ 
Proposed Method & 28.30\%\\
\hline
\end{tabular}
\caption{Results of different method for action detection}
\label{table:4}
\end{table}
\section{Conclusion}
In this paper, we proposed a two-step framework for the task of action detection in aerial images. Since objects in aerial view are represented by a limited number of pixels, it seems necessary to use all those pixels. SSD and other state of the art deep detectors work with input images of relatively small resolutions, therefore, we introduced a framework in which, first we generate high quality object proposals using small images, and then use the real patches in their original sizes for the second step. We proposed a VGG network to extract image features from these proposals and build a latent common layer for the fusion of image and multiple label features. The other important issue is that using multi-label framework can help the network to use the share information between the labels. Results on the Okutama-Action dataset show that the proposed method is more accurate compared to using only SSD for aerial action detection.

\bibliography{Fusion.bib}

\begin{thebibliography}{10}
\providecommand{\url}[1]{#1}
\csname url@samestyle\endcsname
\providecommand{\newblock}{\relax}
\providecommand{\bibinfo}[2]{#2}
\providecommand{\BIBentrySTDinterwordspacing}{\spaceskip=0pt\relax}
\providecommand{\BIBentryALTinterwordstretchfactor}{4}
\providecommand{\BIBentryALTinterwordspacing}{\spaceskip=\fontdimen2\font plus
\BIBentryALTinterwordstretchfactor\fontdimen3\font minus
  \fontdimen4\font\relax}
\providecommand{\BIBforeignlanguage}[2]{{%
\expandafter\ifx\csname l@#1\endcsname\relax
\typeout{** WARNING: IEEEtran.bst: No hyphenation pattern has been}%
\typeout{** loaded for the language `#1'. Using the pattern for}%
\typeout{** the default language instead.}%
\else
\language=\csname l@#1\endcsname
\fi
#2}}
\providecommand{\BIBdecl}{\relax}
\BIBdecl

\bibitem{oren1997pedestrian}
M.~Oren, C.~Papageorgiou, P.~Sinha, E.~Osuna, and T.~Poggio, ``Pedestrian
  detection using wavelet templates,'' in \emph{Computer Vision and Pattern
  Recognition, 1997. Proceedings., 1997 IEEE Computer Society Conference
  on}.\hskip 1em plus 0.5em minus 0.4em\relax IEEE, 1997, pp. 193--199.

\bibitem{tome2016deep}
D.~Tom{\`e}, F.~Monti, L.~Baroffio, L.~Bondi, M.~Tagliasacchi, and S.~Tubaro,
  ``Deep convolutional neural networks for pedestrian detection,'' \emph{Signal
  Processing: Image Communication}, vol.~47, pp. 482--489, 2016.

\bibitem{zhang2016faster}
L.~Zhang, L.~Lin, X.~Liang, and K.~He, ``Is faster r-cnn doing well for
  pedestrian detection?'' in \emph{European Conference on Computer
  Vision}.\hskip 1em plus 0.5em minus 0.4em\relax Springer, 2016, pp. 443--457.

\bibitem{okutama}
M.~Barekatain, M.~Mart{\'\i}, H.-F. Shih, S.~Murray, K.~Nakayama, Y.~Matsuo,
  and H.~Prendinger, ``Okutama-action: An aerial view video dataset for
  concurrent human action detection,'' in \emph{1st Joint BMTT-PETS Workshop on
  Tracking and Surveillance, CVPR}, 2017, pp. 1--8.

\bibitem{faster}
S.~Ren, K.~He, R.~Girshick, and J.~Sun, ``Faster r-cnn: Towards real-time
  object detection with region proposal networks,'' in \emph{Advances in neural
  information processing systems}, 2015, pp. 91--99.

\bibitem{yolo}
J.~Redmon, S.~Divvala, R.~Girshick, and A.~Farhadi, ``You only look once:
  Unified, real-time object detection,'' in \emph{Proceedings of the IEEE
  conference on computer vision and pattern recognition}, 2016, pp. 779--788.

\bibitem{ssd}
W.~Liu, D.~Anguelov, D.~Erhan, C.~Szegedy, S.~Reed, C.-Y. Fu, and A.~C. Berg,
  ``Ssd: Single shot multibox detector,'' in \emph{European conference on
  computer vision}.\hskip 1em plus 0.5em minus 0.4em\relax Springer, 2016, pp.
  21--37.

\bibitem{vgg}
K.~Simonyan and A.~Zisserman, ``Very deep convolutional networks for
  large-scale image recognition,'' \emph{arXiv preprint arXiv:1409.1556}, 2014.

\bibitem{uijlings2013selective}
J.~R. Uijlings, K.~E. Van De~Sande, T.~Gevers, and A.~W. Smeulders, ``Selective
  search for object recognition,'' \emph{International journal of computer
  vision}, vol. 104, no.~2, pp. 154--171, 2013.

\bibitem{rcnn}
R.~Girshick, J.~Donahue, T.~Darrell, and J.~Malik, ``Rich feature hierarchies
  for accurate object detection and semantic segmentation,'' in
  \emph{Proceedings of the IEEE conference on computer vision and pattern
  recognition}, 2014, pp. 580--587.

\bibitem{fastrcnn}
R.~Girshick, ``Fast r-cnn,'' in \emph{Proceedings of the IEEE international
  conference on computer vision}, 2015, pp. 1440--1448.

\bibitem{overfeat}
P.~Sermanet, D.~Eigen, X.~Zhang, M.~Mathieu, R.~Fergus, and Y.~LeCun,
  ``Overfeat: Integrated recognition, localization and detection using
  convolutional networks,'' \emph{arXiv preprint arXiv:1312.6229}, 2013.

\bibitem{everingham2010pascal}
M.~Everingham, L.~Van~Gool, C.~K. Williams, J.~Winn, and A.~Zisserman, ``The
  pascal visual object classes (voc) challenge,'' \emph{International journal
  of computer vision}, vol.~88, no.~2, pp. 303--338, 2010.

\bibitem{deng2009imagenet}
J.~Deng, W.~Dong, R.~Socher, L.-J. Li, K.~Li, and L.~Fei-Fei, ``Imagenet: A
  large-scale hierarchical image database,'' in \emph{Computer Vision and
  Pattern Recognition, 2009. CVPR 2009. IEEE Conference on}.\hskip 1em plus
  0.5em minus 0.4em\relax IEEE, 2009, pp. 248--255.

\bibitem{Read2011}
J.~Read, B.~Pfahringer, G.~Holmes, and E.~Frank, ``Classifier chains for
  multi-label classification,'' \emph{Machine Learning}, vol.~85, no.~3, p.
  333, Jun 2011.

\bibitem{vqa}
S.~Antol, A.~Agrawal, J.~Lu, M.~Mitchell, D.~Batra, C.~Lawrence~Zitnick, and
  D.~Parikh, ``Vqa: Visual question answering,'' in \emph{Proceedings of the
  IEEE International Conference on Computer Vision}, 2015, pp. 2425--2433.

\bibitem{xavier}
X.~Glorot and Y.~Bengio, ``Understanding the difficulty of training deep
  feedforward neural networks,'' in \emph{Proceedings of the Thirteenth
  International Conference on Artificial Intelligence and Statistics}, 2010,
  pp. 249--256.

\end{thebibliography}
\bibliographystyle{IEEEtran}

\end{document}